\documentclass[11pt]{llncs}
\usepackage{times}
\usepackage{url}
\usepackage{latexsym}

\usepackage{graphicx}
\usepackage{float}
\usepackage{caption} 
\title{Extractive Summarization: Limits, Compression, Generalized Model and Heuristics\thanks{Research supported in part by NSF grants DUE 1241772, CNS 1319212 and DGE 1433817}}

\author{Rakesh Verma and Daniel Lee}
\institute{Computer Science Department  \\
  University of Houston  \\
  Houston, TX 77204, USA\\
  {\tt rmverma@cs.uh.edu},  {\tt dljr0122@cs.uh.edu}
}

\date{}

\begin{document}
\maketitle
\begin{abstract}
  Due to its promise to alleviate information overload, text summarization has attracted the attention of many  researchers. However, it has remained a serious challenge. Here, we first prove empirical limits on the recall (and F1-scores) of extractive summarizers on the DUC datasets under ROUGE evaluation for both the single-document and multi-document summarization tasks. Next we define the concept of compressibility of a document and present a new model of  summarization, which generalizes existing models in the literature and integrates several dimensions of the summarization, viz., abstractive versus extractive, single versus multi-document, and syntactic versus semantic. Finally, we examine some new and existing single-document summarization algorithms in a single framework and compare with state of the art summarizers on DUC data.  
  
\end{abstract}

\section{Introduction}

Automatic text summarization is the holy grail for people battling information overload, which becomes more and more acute over time. Hence it has attracted many researchers from diverse fields since the 1950s. However, it has remained a serious challenge, especially in the case of single news articles. The single document summarization competition at Document Understanding Conferences (DUC) was abandoned after only two years, 2001-2002, since many automatic summarizers could not outperform a baseline summary consisting of the first 100 words of a news article. Those that did outperform the baseline could not do so in a statistically significant way \cite{nenkova05}. Summarization can be extractive or abstractive \cite{mani:99}: in extractive summarization sentences are chosen from the article(s) given as input, whereas in abstractive summarization sentences may be generated or a new representation of the article(s) may be output. 

Extractive summarization is popular, so we explore whether there are inherent limits on the performance of such systems.\footnote{Surprisingly, despite all the attention extractive summarization has received, to our knowledge, no one has explored this question before.} We then generalize existing models for summarization and define compressibility of a document. We explore this concept for documents from three genres and then unify new and existing heuristics for summarization in a single framework. Our contributions:  
\begin{enumerate}
	\item We show the limitations of single and multi-document {\em extractive summarization} when the comparison	is with respect	to gold-standard human-constructed abstractive summaries on DUC data (Section~\ref{sec-limit}). 
		\begin{enumerate}
        	\item Specifically, we show that when the documents themselves from the DUC 2001-2002 datasets are compared using ROUGE \cite{lin-hovy:03} to abstractive summaries, the average Rouge-1 (unigram) recall is around 90\%. On ROUGE evaluations, no extractive summarizer can do better than just returning the document itself (in practice it will do much worse because of the size constraint on summaries). 
			\item For multi-document summarization,  we show limits in two ways: (i) we concatenate the documents in each set and examine how this ``superdocument" performs as a summary with respect to the manual abstractive summaries, and (ii) we study how each document measures up against the manual summaries and then average the performance of all the documents in each set. 
   		\end{enumerate}
    \item Inspired by this view of documents as summaries, we introduce and explore a
	generalized model of summarization (Section \ref{summ_model}) that unifies the three different dimensions:
	abstractive versus extractive, single versus multi-document and syntactic versus semantic. 
	\begin{enumerate}
		\item We prove (in Appendix) that constructing certain extractive summaries 
		is isomorphic to the min cover problem for sets, which shows that
		not only is the optimal summary problem NP-complete but it has a
		greedy heuristic that gives a multiplicative logarithmic approximation.  
        \item Based on our model, we can define the {\em compressibility} of a
		document. We study this notion for different genres of articles
		including: news articles, scientific articles and short stories. 
		\item We present new and existing heuristics for single-document summarization, which represent different time and compressibility trade-offs.  
        We compare them against existing summarizers proven on DUC datasets.
	\end{enumerate}
\end{enumerate}
Although many metrics have been proposed (more in Section \ref{sec-related}), we use ROUGE because of its popularity, ease of use and correlation with human evaluations. 

\section{Related Work} \label{sec-related}
Most of the summarization literature focuses on single-document and multi-document summarization algorithms and frameworks 
rather than limits on the performance of summarization systems. As pointed out by
\cite{dangO08}, competitive summarization systems are
typically extractive, selecting representative sentences,
concatenating them and often compressing them to squeeze in
more sentences within the constraint. The summarization literature is vast, so we refer the reader to the recent survey \cite{gambhirG17}, which is fairly comprehensive for summarization research until 2015. Here, we give a sampling of the literature and focus more on recent research and/or evaluation work. 

{\bf Single-document extractive summarization.} For {\em single-document summarization}, \cite{martinsS09} explicitly model extraction and compression,
but their results showed a wide variation on a subset of 140 documents from the DUC 2002 dataset, and \cite{parveenRS15} focused on topic coherence with a graphical structure with separate importance, coherence and topic coverage functions. In \cite{parveenRS15}, the authors present results for single-document summarization on a subset of PLOS Medicine articles and DUC 2002 dataset without mentioning the number of articles used. An algorithm combining syntactic and semantic features was presented by \cite{rvcicling}, and graph-based summarization methods in \cite{vanderwendeAO04,erkanR04,mihalceaT04,kumarKV13}. Several systems were compared against a newly-devised supervised method on a dataset from Yahoo in \cite{mehdadSTR16}. 

{\bf Multi-document extractive summarization.} For {\em multi-document summarization},
extraction and redundancy/compression of sentences have been modeled  by integer linear programming and approximation algorithms \cite{mcdonald07,gillickF09,berg-kAO11,almeidaM13,liAO14,boudinMF15,yogatamaLS15}. Supervised and semi-supervised learning based extractive summarization was studied in \cite{wongWL08}. Of course, single-document summarization can be considered as a special case, but {\em no} experimental results are presented for this important special case in the papers cited in this paragraph.   

{\bf Abstractive summarization.} Abstractive summarization systems include
\cite{careniniC08,ganesanAO10,cheungP14,liuAO15,rushCW15,chopraAR16}.  

{\bf Frameworks.} Frameworks for single-document summarization were presented in \cite{filatova04,mcdonald07,takamuraO09}, and some multi-document summarization frameworks are in  \cite{hiraoYMNM13,yoshidaSHN14}.

{\bf Metrics and Evaluation.} Of course, ROUGE is not the only metric for evaluating summaries. Human evaluators were used at NIST for scoring summaries on seven different metrics such as linguistic quality, etc. There is also the Pyramid approach \cite{passonneauCGP13} and BE \cite{tratzH08}, for example. Our choice of ROUGE is based on its popularity, ease of use, and correlation with human assessment \cite{lin-hovy:03}. Our choice of ROUGE configurations includes the one that was found to be best according to the paper \cite{graham15}.
   
\section{Limits on Extractive Summarization} \label{sec-limit}
In all instances the ROUGE evaluations include the best schemes as shown by \cite{graham15}, which are usually Rouge-2 (bigram) and Rouge-3 (trigram) with stemming and stopword elimination. We also include the results without stopword elimination. The only modification was if the original parameters limited the size of the generated summary; we removed that option. 
\subsection{Single-document Summarization} 
To study limits on extractive summarization, we will pretend that the document is itself a summary that needs to be evaluated against the human (abstractive) summaries created by NIST experts. Of course, the ``precision" of such a summary will be very low, so we focus on recall (and F-score by letting the document get all its recall from the same size as the human summary (100 words)). Table~\ref{tab-duc02} shows that, for the DUC 2002\footnote{2002 was the last year in which the single document summarization competition was	held by NIST.} dataset, when the {\em document themselves} are considered as summaries and evaluated against a set of 100-word human abstractive summaries, the average Rouge-1 (unigram) \cite{lin-hovy:03} score is approximately 91~\%. 
Tables~\ref{tab-duc01} through \ref{tab-duc05} and Figures~\ref{avg04} and \ref{avg05} use the following abbreviations: (i) R-n means ROUGE metric using n-gram matching, and (ii) lowercase $s$ denotes the use of stopword removal option.

\begin{table*}
	\centering
		\begin{minipage}[t]{.4\textwidth}
			\centering\setlength{\tabcolsep}{6pt}
			\begin{tabular}{l c c c}
				\hline		
				\textbf{Metric} & \textbf{$\mu$} &  \textbf{$\sigma$} & \textbf{Range} \\
				\hline
				R-1 & 0.909 & 0.069 & 0.49-1.00 \\ 
				R-1s & 0.879 & 0.103 & 0.15-1.00 \\ 
				R-2 & 0.555 & 0.167 & 0.06-0.96 \\ 
				R-2s & 0.505 & 0.179 & 0.02-0.95 \\ 
				R-3 & 0.376 & 0.192 & 0.01-0.93 \\ 
				R-3s & 0.315 & 0.189 & 0.00-0.89 \\ 
				R-4 & 0.278 & 0.190 & 0.00-0.90 \\ 
				R-4s & 0.213 & 0.175 & 0.00-0.84 \\ 
				\hline				
			\end{tabular}
			\caption{Rouge Recall on DUC 2001, Document as summary.} \label{tab-duc01} 
		\end{minipage}
		\hspace{.1\textwidth}
		\begin{minipage}[t]{.4\textwidth}
			\centering\setlength{\tabcolsep}{6pt}
			\begin{tabular}{l c c c}
				\hline		
				\textbf{Metric} & \textbf{$\mu$} &  \textbf{$\sigma$} & \textbf{Range} \\
				\hline
				R-1 & 0.907 & 0.045 & 0.57-1.00 \\
				R-1s & 0.889 & 0.059 & 0.64-1.00 \\
				R-2 & 0.555 & 0.111 & 0.22-0.85 \\
				R-2s & 0.509 & 0.117 & 0.21-0.87 \\
				R-3 & 0.372 & 0.124 & 0.04-0.75 \\
				R-3s & 0.311 & 0.123 & 0.04-0.76 \\
				R-4 & 0.272 & 0.118 & 0.01-0.67 \\
				R-4s & 0.204 & 0.112 & 0.01-0.68 \\ 
				\hline
			\end{tabular}
			\caption{Rouge Recall on DUC 2002, Document as summary.} 
			\label{tab-duc02} 
		\end{minipage}
\end{table*}

This means that on the average about 9\% of the words in the human abstractive summaries {\em do not appear in the documents}. Since extractive automatic summarizers extract all the sentences from the documents given to them for summarization, clearly no extractive summarizer can have Rouge-1 recall score higher than the documents themselves on any dataset, and, in general, the recall score will be lower since the summaries are limited to 100 words whereas the documents themselves can be arbitrarily long. Thus, we establish a limit on the Rouge recall scores for extractive summarizers on the DUC datasets. The DUC 2002 dataset has 533 {\em unique} documents and most include two 100-word human abstractive summaries.
We note that if extractive summaries are also exactly 100 words each, then the precision can also be no higher than recall score. In addition, since the F1-score is upper bounded by the highest possible recall score. Therefore in the single document summarization, no extractive summarizer can have an average F1-score better than about 91\%. When considered in this light, the best current extractive single-document summarizers achieve about 54\% of this limit on DUC datasets, e.g., see \cite{rvcicling,kumarKV13}.


\subsubsection{ROUGE insights} In Table~\ref{tab-duc02}, comparing R-1 and R-1s, we can see an increase in the lower range of recall values with stopword removal. This occurred with Document \#250 (App.~\ref{doc250}). Upon deeper analysis of ROUGE, we found that it does not remove numbers under stopword removal option. Document \#250 had a table with several numbers. In addition ROUGE treats numbers with the comma character (and also decimals such as 7.3) as two different numbers (e.g. 50,000 become 50 and 000). This boosted the recall because after stopword removal, the summaries significantly decreased in unigram count, whereas the overlapping unigrams between document and summary did not drop as much. Another discovery is that documents with long descriptive explanations end up with lower recall values with stopword removal. Tabel~\ref{tab-duc01} shows a steep drop on the lower range values from R-1 to R-1s. When looking at the lower scoring documents, the documents usually had explanations about events, whereas the summary skipped these explanations.

\subsection{Multi-document Extractive Summarization}
For multi-document summarization, there are at least two different scenarios in which to explore limits on extractive summarization. The first is where documents belonging to the same topic are concatenated together into one super-document and then it is treated as a summary. In the second, we compare each document as a summary with respect to the model summaries and then average the results for documents belonging to the same topic. 

For multi-document summarization, experiments were done on data from DUC datasets for 2004 and 2005. The data was grouped into document clusters. Each cluster held documents that were about a single topic. For the 2004 competition (DUC 2004), we focused on the English document clusters. There were a total of 50 document clusters and each document cluster had an average of 10 documents. DUC 2005 also had 50 documents clusters, however, there were a minimum of 25 documents for each set.

Please note that since the scores for R-3 and R-4 were quite low (best being 0.23) these scores are not reported here.

\subsubsection{Super-document Approach}
Now we consider the overlap between the documents of a cluster with the human summaries of those clusters. So for this limit on recall, we create \textbf{super-documents}. Each super-document is the concatenation of all the documents for a given document set. These super-documents are then evaluated with ROUGE against the model human summaries. Any extractive summary is limited to only these words, so the recall of a perfect extractive system can only reach this limit. The results can be seen in Table~\ref{tab-duc04} and Table~\ref{tab-duc05}.

\begin{table}
	\centering
	\begin{minipage}[t]{.4\textwidth}\centering\setlength{\tabcolsep}{6pt}
	\begin{tabular}{l c c c}
		\hline
		\textbf{Metric} & \textbf{$\mu$} &  \textbf{$\sigma$} & \textbf{Range} \\
		\hline
		R-1 & 0.938 & 0.021 & 0.89-0.97 \\ 
        R-1s & 0.904 & 0.030 & 0.82-0.96 \\ 
        R-2 & 0.474 & 0.057 & 0.36-0.60 \\ 
        R-2s & 0.351 & 0.061 & 0.22-0.48 \\
        \hline
	\end{tabular}
	\caption{ROUGE Recall on DUC 2004, Super-document as summary.}
	\label{tab-duc04}
	\end{minipage}
	\hspace{.1\textwidth}
	\begin{minipage}[t]{.4\textwidth}\centering\setlength{\tabcolsep}{6pt}
	\begin{tabular}{c c c c}

		\hline
		\textbf{Metric} & \textbf{$\mu$} &  \textbf{$\sigma$} & \textbf{Range} \\
		\hline
        R-1 & 0.969 & 0.018 & 0.88-0.99 \\
        R-1s & 0.949 & 0.029 & 0.81-0.99 \\
        R-2 & 0.537 & 0.080 & 0.30-0.73 \\ 
        R-2s & 0.396 & 0.087 & 0.18-0.64 \\ 
        \hline
	\end{tabular}
	\caption{ROUGE Recall on DUC 2005, Super-document as summary.}
	\label{tab-duc05}
	\end{minipage}%
\end{table}
\subsubsection{Averaging Results of Individual Documents}
\begin{figure*}
\centering
  \begin{minipage}[t]{.45\textwidth}
  \centering
  \includegraphics[width=1\textwidth]{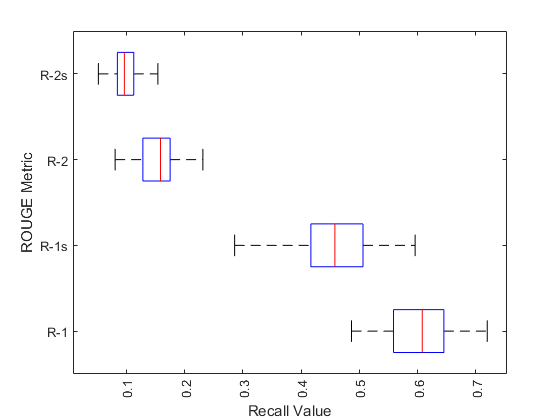}
  \caption{Distribution of Avg for DUC 2004}
  \label{avg04}
  \end{minipage}
  \hfill
  \begin{minipage}[t]{.45\textwidth}
  \centering
  \includegraphics[width=1\textwidth]{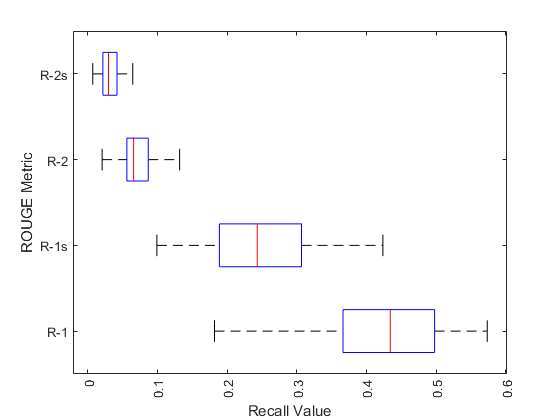}
  \caption{Distribution of Avg for DUC 2005}
  \label{avg05}
  \end{minipage}
\end{figure*}

Here we show a different perspective on the upper limit of extractive systems. We treat each document as a summary to compare against the human summaries. Since all the documents are articles related to a specific topic, these documents can be viewed as a standalone perspective. For this experiment we obtained the ROUGE recall of each document  and then averaged them for each cluster. The distribution of the averages are presented in Figure~\ref{avg04} and Figure~\ref{avg05}. Here the best distribution average is only about 60\% and 42\% for DUC 2004 and DUC 2005, respectively. The best system did approximated 38\% in DUC 2004 and 46\% in DUC 2005

\section{A General Model for Summarization}\label{summ_model}
Now we introduce our model and study its implications. Consider the process of human summarization. 
The starting point is a document,  which contains a sequence of sentences that in
turn are sequences of words. However, when a human is given a document
to summarize, the human does not choose full sentences to extract from
the document like extractive summarizers. Rather, the human
first tries to understand the document, i.e., builds an abstract
mental representation of it, and then writes a summary of the document 
based on this.  

Therefore, we formulate a model for {\em semantic summarization} in the abstract world of
thought units,\footnote{we prefer thought units because a sentence is defined as a complete thought} which can be specialized to syntactic summarization by using words in place of thought units. We hypothesize that a document is a collection of
thought units, some of which are more important than others, with a
mapping of sentences to thought units. The natural mapping is that of
implication or inclusion, but this could be partial implication, not necessarily full
implication. That is, the mapping could associate a degree to represent
that the sentence only includes the thought unit partially. 
A summary
must be constructed from sentences, {\em not necessarily in the document}, that cover as many 
of the important thought units as possible, i.e., maximize the
importance score of the thought units selected, within a given size
constraint $C$. We now define it formally for single and multi-document summarization. {\em Our model can naturally represent abstractive versus extractive dimension of summarization.} 

Let $S$ denote an infinite set of
sentences, $T$ an infinite set of thought units, and $I: S \times T
\rightarrow R$ be a mapping that associates a non-negative real number
for each sentence $s$ and thought unit $t$ that measures the degree to
which the thought unit is implied by the sentence $s$. Given a
document $D$, which is a finite sequence of sentences from $S$, let $S(D) \subset S$ be the finite set of sentences in $D$ and $T(D)
\subset T$ be the finite set of thought units of $D$. Once thoughts are assembled
into sentences in a document with its sequencing (a train of thought) and title(s), this
imposes a certain  ordering\footnote{Since some thought units in the same sentence } of importance on these thought
units, which is denoted by a scoring function $W_D: T \rightarrow
R$. The size of a document is denoted by $|D|$, which could be, for example, the
total number of words or sentences in the document. A  size
constraint, $C$, for the summary, is  a function of $|D|$, e.g., a
percentage of $|D|$, or a fixed number of words or sentences in which case 
it is a constant function. A summary of $D$, denoted by $summ(D)
\subset S$, is a finite sequence of sentences that
attempts to represent the thought units of $D$ as best as possible
within the constraint $C$. 
The size of a summary, $|summ(D)|$ is
measured using the same procedure for measuring $|D|$. With
	these notations, for each thought unit $t \in T(D)$, we define
	the score assigned to $summ(D)$ for expressing thought unit
	$t$ as $Ts(t, summ(D)) = max \{{\cal I}(s, t) ~|~ s \in
	summ(D) \}$. Formally, the summarization problem is then,
	select $summ(D)$ to maximize $Utility(summ(D))$:  
	
	\[   \sum_{t \in T(D)} W_D(t)*Ts(t,
	summ(D)) \] subject to the constraint $|summ(D)| \leq C$.

Note that our model
can represent some aspects of {\em summary coherence} as well by imposing
the constraint that the sequencing of thought units in the summary be
consistent with the ordering of thought units in the document. 

For the {\bf multi-document case}, we are given a 
$Corpus = \{D_1, D_2, \ldots D_n\}$, each $D_i$ has its own sequencing of
sentences and thought units, which could conflict with other documents. One must
resolve the conflicts somehow when constructing a single summary of
the corpus. Thus, for multi-document summarization, we
hypothesize that $W_{Corpus}$ is a total ordering that is maximally
consistent with the $W_{D_i}$'s by which we mean that if two thought
units are assigned the same relative importance by every document in
the collection that includes them, then the same relative order is
imposed by $W$ as well, otherwise $W$ chooses a relative order that is
best represented by the collection and this could be based on a
majority of the documents or in other ways. With this, our previous
definition extends to multi-document summarization as well, but we
replace $summ(D)$ by $summ(Corpus)$, $W_D$ by
$W_{Corpus}$, and $T(D)$ by $T(Corpus)$. In the multi-document case, the summary coherence can be
defined as the constraint that the sequencing of thought units in a
summary be maximally consistent with the sequencing of thought units
in the documents and in conflicting cases makes the same choices as
implied by $W_{Corpus}$. 

The function $W$ is a crucial ingredient that allows us
to capture the sequencing chosen by the author(s) of the document(s),
without $W$ we would get the bag of words models popular in previous
work. We note that $W$ does need to respect the sequencing in the
sense that it is not required to be a decreasing (or even
non-increasing) function with sequence position. This flexibility is
needed since $W$ must fit the document structure. 

As defined our model covers {\em abstractive summarization}
directly since it is based on sentences that are not restricted to
those within $D$. For {\em extractive summarization}, we need to impose the
additional constraint $summ(D) \subseteq S(D)$ for single-document, and $summ(D) \subseteq S(Corpus)$, where $S(Corpus) = \cup_i S(D_i)$, for multi-document summarization. 
Some other important special cases of our model as as follows:
\begin{enumerate}
\item Restricting $I(S,T)$ to a boolean-valued function. This gives rise to the
``membership'' model and avoids partial membership. 
\item Restricting $W_D(t)$ to a constant function. This would give
  rise to a ``bag of thought units'' model and it would treat all
  thought units the same. 
\item Further, if ``thought units'' are limited to be all words, or all words
  minus stopwords, or key phrases of the document, and under extractive constraint, we get previous
  models of \cite{filatova04,mcdonald07,takamuraO09}. This also means that the
  optimization problem of our model is NP-hard at least and
  NP-complete when $W_D(t)$ is a constant function and $I(S,T)$ is
  boolean-valued. 
\end{enumerate}
\begin{theorem}

The optimization problem of the model is at least NP-hard. It is
NP-complete when $I(S,T)$ is boolean-valued, $W_D(t)$ is a constant
function and thought units are: words, or all words minus stopwords or
key phrases of the document, with sentence size and summary size constraint being measured in these same syntactic units. We call these NP-complete cases extractive coverage summarization collectively. 
\end{theorem}
{\em Proof:} Reduction from the set cover problem - proof in Appendix. \medskip

Based on this generalized model, we can define: 
\begin{definition}
	The {\em extractive compressibility} of a document is the smallest
	size collection of sentences from the document that cover its thought units. If the
	thought units are words, we call it  the {\em word extractive
	compressibility}. 
\end{definition}
\begin{definition}
	The {\em abstractive compressibility} of a document is the smallest size
	collection of arbitrary sentences that cover its thought units. If the
	thought units are words, we call it the {\em word abstractive
	compressibility}. 
\end{definition}
\begin{definition}
	The {\em compression rate} or {\em incompressibility} of a document is defined as $\kappa/N$, where $\kappa$ is the size of the compressibility of the document, and $N$ is the original size of the document.
    
\end{definition}
Similarly, we can define corresponding compressibility notions for key phrases, words
minus stopwords, and thought units.

\begin{figure}[h]
\centering
	\caption{Imcompressibility vs. Sentence Count}
	\includegraphics[width=.7\textwidth]{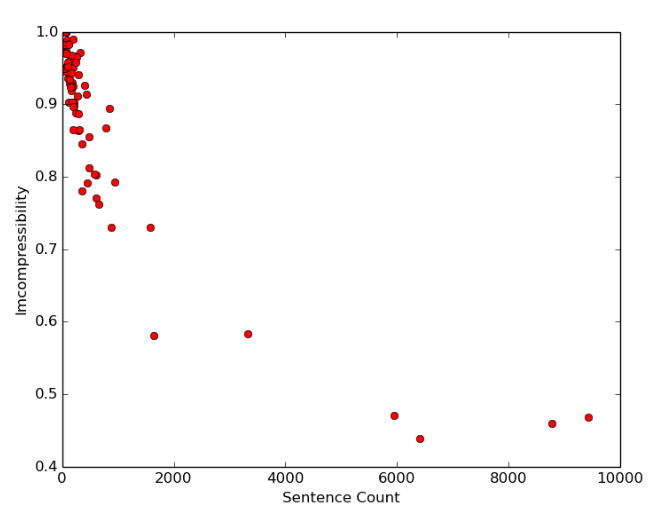} \label{vs_count}
\end{figure}

We investigate compressibility of three different genres: news articles, scientific papers and short studies. For this purpose, 
25 news articles, 25 scientific
papers, and 25 short stories were collected. The 25 news articles were randomly selected from several sources and covered disasters, disaster
recovery, prevention, and critical infrastructures. Five scientific
papers, on each of the following five topics: cancer research,
nanotechnology, physics, NLP and security, were chosen at random. Five short
stories each by Cather, Crane, Chekhov, Kate Chopin, and O'Henry were
randomly selected. Experiments showed that large sentence counts lead to decrease imcompressibility. Figure~\ref{vs_count} shows a direct relationship between document size and incompressibility.


\subsection{Algorithms for Single-document Summarization}	
	We have implemented several new and existing heuristcs in a tool called DocSumm written in Python. Many of our heuristics  revolve around the TF/IDF ranking, which has been shown to do well in tasks involving summarization. 
	
	TF/IDF ranks the importance of words across a corpus. This ranking system was compared to other popular keyword	identification algorithms and was found to be quite competitive in results {\cite{Meseure13}}. In this paper, the authors  compared TextRank, SingleRank, ExpandRank, KeyCluster, Latent Semantics Analysis, Latent Dirichlet Analysis and TF/IDF. $N$ keywords, where $N$ varied from 5 to 100 in steps of 5, from the DUC documents were extracted using each algorithm and the F1-score was calculated using human summaries as models. The experiments showed that TF/IDF consistently performs as well if not better than other algorithms. To apply to the domain of single-document summarization, we define a corpus as the document itself. The documents referred to in inverse document frequency are the individual sentences and the terms remain the same, words. The value of a sentence is then the sum of the TF/IDF scores of the words in the sentence.
	
	DocSumm, includes both greedy and dynamic programming based algorithms. The greedy algorithms use the chosen scoring metric to evaluate every sentence of a document. It then simply selects the highest scoring sentence, until either a given threshold of words are met or every word is covered in the document. Besides the choices for the scoring metrics, there are several other options (normalization of weights, stemming, etc.) that can be toggled for evaluation. Appendix~\ref{apx:tool} gives a brief description of those options.
	
	DocSumm includes two dynamic programming algorithms. One provides an optimal solution, i.e., the minimum number sentences necessary to cover all words of the document. This can be viewed as the bound on maximum compression of a document for extractive summary. This algorithm is a bottom-up approach that builds set covers of all subsets of the original document's thought units (i.e. words for our experiments), beginning with the smallest unit, a single word. We did implement a top-down version based on recursion, but this algorithm quickly runs out of time/space because of repeated computations.
    
    In addition to this optimal algorithm, DocSumm also implements a version of the algorithm presented in \cite{mcdonald07}. McDonald frames the problem of document summarization as the maximization of a scoring function that is based on relevance and redundancy. In essence, selected sentences are scored higher for relevance and scored lower for redundancy. If the sentences of a document are considered on a inclusion/exclusion basis, then the problem of document summarization reduces to the 0-1 Knapsack problem. However, McDonald's algorithm is approximate, because the inclusion/exclusion of the algorithm influences the score of other sentences. A couple of greedy algorithms and a dynamic programming algorithm of DocSumm appeared in \cite{takamuraO09}, the rest are new.

\subsection{Results}
Our results include experiments on running time comparisons of DocSumm's algorithms. In addition we compare the performance measures of DocSumm on DUC 2001 and DUC 2002 datasets.
	
\noindent {\bf Run times} The dataset for running times is created by sampling sentences from the book of Genesis. We created documents of increasing lengths, where length is measured in verses. The verse count ranged from 4 to 320. However, for documents greater than 20 sentences, the top-down dynamic algorithm runs out of memory. So there are no results on the top-down exhaustive algorithm. Table~\ref{table timetest} shows slight increases in time as the document size increase. For both tfidf and bottom-up there is a significant increase in running time. 
	
	\begin{table*}
		\small
		\centering\setlength{\tabcolsep}{6pt}
		\begin{tabular}{ c c c c c c c c c c}
			\hline
			\vtop{\hbox{\strut \bf verse}
				\hbox{\strut \bf count}}&
			\vtop{\hbox{\strut \bf greedy}
				\hbox{\strut \bf size}}&
			\vtop{\hbox{\strut \bf greedy}
				\hbox{\strut \bf size+d}}&
			\vtop{\hbox{\strut \bf greedy}
				\hbox{\strut \bf tfidf}}&
			\vtop{\hbox{\strut \bf bottom-}
				\hbox{\strut \bf up}}\\
			\hline
			4	& 33	& 32	& 32	
			& 34\\
			8	& 32	& 32	& 33	
			& 36\\
			12	& 33	& 33	& 35	
			& 36\\
			16	& 35	& 35	& 36	
			& 39\\
			20	& 35	& 35	& 37	
            & 38\\
			40	& 43	& 43	& 48	
			& 70\\
			80	& 59	& 57	& 101	
            & 101\\
			160	& 92	& 90	& 331	
            & 408\\
			320	& 170	& 167	& 1520	
			& 1708\\
			\hline
		\end{tabular}
		\caption{\label{table timetest} Running Times of algorithms in milliseconds.}
	\end{table*}
\subsubsection{Summarization}
We now compare the heuristics for single-document summarization on DUC 2001 and DUC 2002 dataset. For the 305 unique documents of the DUC 2001 dataset we compared the summaries of DocSumm algorithms. The results were in line with the analysis of the three domains. 
	
	For each algorithm, we truncated the solution set as soon as a threshold of 100 words was covered. The ROUGE scores of the algorithms were in line with the compressibility performances. The size algorithms performed similarly and the best was the bottom-up algorithm with ROUGE F1 scores of 0.444, 0.273 and 0.408 for ROUGE-1, ROUGE-2 and ROUGE-LCS, respectively. The tfidf algorithm performance was not significantly different.
	
\subsubsection{Comparison} On the 533 {\em unique} articles in the DUC 2002 dataset, we now compare our greedy and dynamic solutions against the following
	classes of systems:
	(i) two top of the
	line single-document summarizers, SynSem \cite{rvcicling}, and the best extractive
	summarizer from \cite{kumarKV13}, which we call KKV, (ii) top five (out of 13) systems, S28, S19,
	S29, S21, S23,  from DUC 2002
	competition, (iii) TextRank, (iv) MEAD, (v) McDonald Algorithm and (vi) the DUC 2002 Baseline summaries consisting of the
	first 100 words of news articles. The Baseline did very well in the
	DUC 2002 competition - only two out of 13 systems, S28 and S19,  managed to get a
	higher F1 score than the Baseline. For this comparison, all manual abstracts and system summaries are truncated to exactly
	100 words whenever they exceed this limit.
\begin{table}[t]
	\centering\setlength{\tabcolsep}{6pt}
	\begin{tabular}{c c c c c}
		\hline
		\bf Algorithm & \bf ROUGE-1 & \bf ROUGE-2 & \bf ROUGE-LCS \\
		\hline
		size & 0.430 & 0.262 & 0.295 \\
		
		size+d & 0.433 & 0.265 & 0.398 \\
		tfidf & 0.440 & 0.272 & 0.406 \\
		bottom-up & 0.444 & 0.273 & 0.408 \\
		mcdonald & 0.428 & 0.254 & 0.387 \\
		MEAD & 0.447 & 0.210 & 0.298 \\ 
		TextRank & 0.446 & 0.208 &0.288 \\
		SynSem & 0.465 & N/A & N/A \\
		KKV & 0.490* & 0.228* & N/A\\
		S23 & 0.450 & 0.218 & 0.299\\
		S29 & 0.453 & 0.212 & 0.300 \\
		S21 & 0.460 & 0.219 & 0.305 \\
		Baseline  & 0.462& 0.222& 0.301\\
		S19 & 0.463 & 0.226 & 0.312\\
		S28 & 0.467 & 0.227 & 0.309\\
		\hline
	\end{tabular}
	\caption{\label{rouge_scores} F1 scores on 100  word summaries for DUC 2002 documents}
\end{table}
		
	Note that the results for SynSem are from
	\cite{rvcicling}, who also used only the 533 {\em
		unique} articles in the DUC 2002 dataset. Unfortunately, the
	authors did not report the Rouge bigram
	(ROUGE-2)  and Rouge LCS (ROUGE-L) 
	F1 scores in  \cite{rvcicling}. KKV's results are
	from \cite{kumarKV13}, who did {\em not} remove the 33 duplicate
	articles in the DUC 2002 dataset,  which is why
	we flagged those entries in Table~\ref{rouge_scores} with *. Hence their
	results are not comparable to
	ours. In addition KKV did not report ROUGE-LCS scores.
	We observe that for Rouge unigram (ROUGE-1) F1-scores the dynamic	optimal algorithm performs the best amongst the algorithms of DocSumm.  However, it still falls behind the Baseline. When we consider Rouge bigram (ROUGE-2) F1-scores Dynamic and Greedy outperform the rest of the
	field (surprisingly even \cite{kumarKV13}). The margin of out-performance is even more pronounced in ROUGE-LCS F1-scores.

\section{Conclusions and Future Work}
We have shown limits on the recall of automatic extractive summarization on DUC datasets under ROUGE evaluations. Our limits show that the current state-of-the-art systems evaluated on DUC data \cite{rvcicling,kumarKV13} are achieving about 54\% of this limit (Rouge-1 recall) for single-document summarization and the best systems for multi-document summarization are achieving only about a third of their limit. This is encouraging news, but at the same time there is much work remaining to be done on summarization. We also explored compressibility, a generalized model, and new and existing heuristics for single-document summarization.

To our knowledge, compressibility the way we have defined and studied it is a new concept and we plan to investigate it further in future work. We believe that compressibility could prove to be a useful measure to study the performance of automatic summarization systems and also perhaps for authorship
detection if, for instance, authors are shown to be consistently compressible. 
\section*{Acknowledgments} We thank the reviewers of CICLING 2017 for their constructive comments. 

	\bibliography{sac,summarization}
	\bibliographystyle{splncs03}

\newpage
\clearpage
\appendix
\section{Appendix - Proof of Theorem 1}
	
 Reduction from the set cover problem for
NP-hardness.  Given a universe U, and a family of S of subsets
of U, a {\em cover} is a subfamily C of S whose union is U. In
the set cover problem the input is a pair (U,S) and a number
$k$, the question is whether there is a cover of size at most
$k$. We reduce set cover to summarization as follows. For each
member $u$ of $U$, we select a thought unit $t$ from ${\cal
	T}$ and a clause $c$ that expresses $t$. For each set $S$ in
the family, we construct a sentence $s$ that consists of the
clauses corresponding to the members of $S$ (${\cal I}$ is
boolean-valued). We assemble all the sentences into a
document. The capacity constraint $C = k$ and represents the
number of sentences that we can select for the summary. It is
easy to see that a cover corresponds to a summary that
maximizes the Utility and satisfies the capacity constraint
and vice versa.   \qed  

Of course, the document constructed above could be somewhat repetitive, but even ``real'' single documents do have some redundancy. Connectivity of clauses appearing in the same sentence can be ensured by choosing them to be facts about a person's life for example. We call the NP-complete cases of the theorem, extractive coverage summarization collectively. For this case, it is easy to design a greedy strategy that gives a logarithmic approximation ratio \cite{hochbaum82} and an optimal dynamic programming one that is exponential in the worst case. 

\section{Appendix - DocSumm Tool}\label{apx:tool}
	\begin{table}
		\small
		\centering\setlength{\tabcolsep}{6pt}
		\begin{tabular}{l l}
			\hline
			\bf Option & \bf Description \\ \hline
			-c size & scoring based on lenght of sentence \\
			-c tfidf & tf based on whole document, idf based on whole document \\
			-w, -{}-stopword & removes stopwords\\ 
			-s, -{}-stem & applies stemming to words\\ 
			-d, -{}-distinct &  removes duplicate words per sentence\\
			-n, -{}-normalize & normalizes scores by sentence word count\\
			-u, -{}-update & updates scores after each greedy selection\\
			-e, -{}-echo & enables summary mode \\
			-t, -{}-threshold & sets the number of words in summary \\ \hline
		\end{tabular}
		\caption{\label{options} Options for DocSumm tool.}
	\end{table}

\clearpage
\section{Appendix - Document 250, AP900625-0153, from DUC 2002}\label{doc250}
\begin{figure}
  \centering
  \includegraphics[width=.75\textwidth]{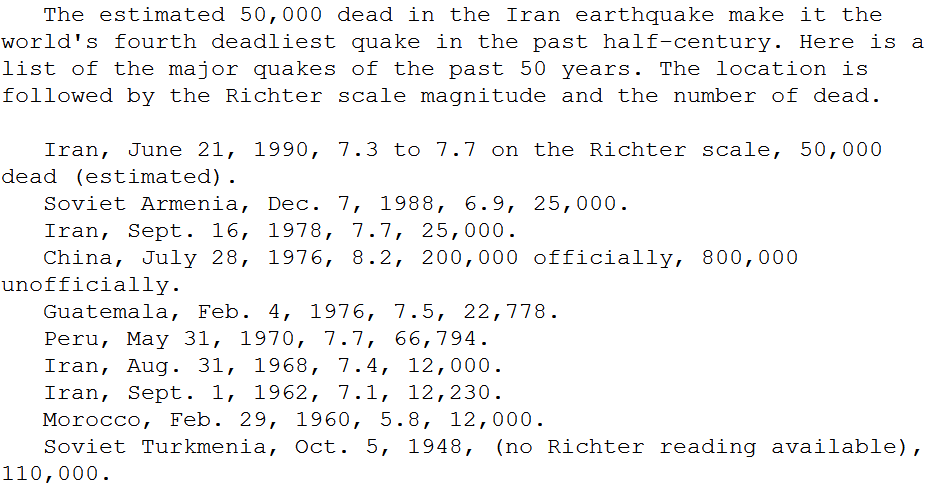}
  \caption{Original Document}
  \label{fig:ori}
\end{figure}
\begin{figure}
  \centering
  \includegraphics[width=1\textwidth]{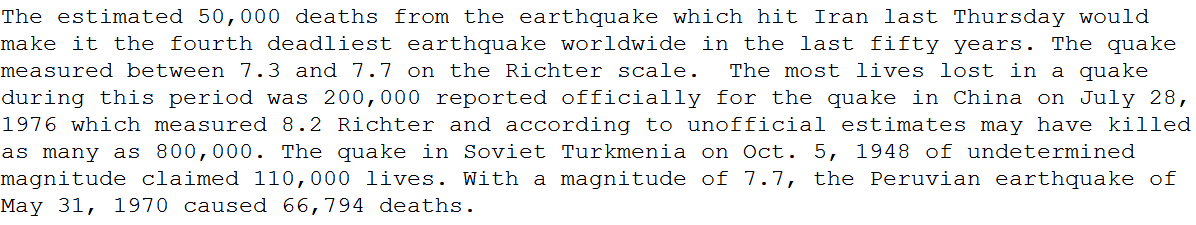}
  \caption{Model Summary 1}
  \label{fig:summ1}
\end{figure}
\begin{figure}
  \centering
  \includegraphics[width=1\textwidth]{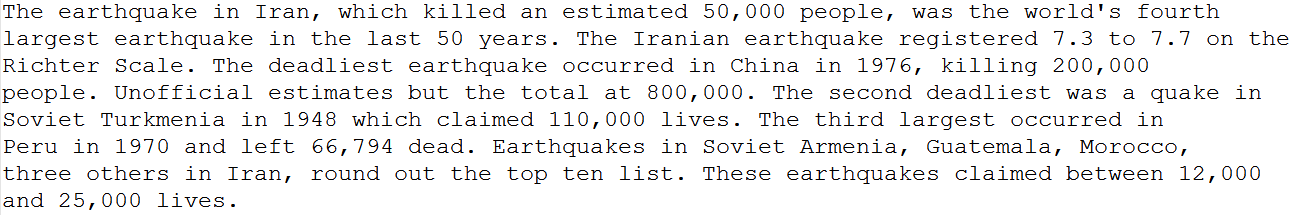}
  \caption{Model Summary 2}
  \label{fig:summ2}
\end{figure}
\end{document}